\documentclass[letterpaper, 10 pt, conference]{ieeeconf}  

\IEEEoverridecommandlockouts                              

\usepackage{graphics} 
\usepackage{epsfig} 
\usepackage{mathptmx} 
\usepackage{times} 
\usepackage{amsmath} 
\usepackage{amssymb}  
\usepackage{xcolor}
\usepackage{url}
\usepackage{graphicx}
\usepackage[ruled,vlined]{algorithm2e}
\usepackage{float}
\usepackage{gensymb}
\usepackage{xcolor}
\usepackage{pifont}
\usepackage{diagbox}
\usepackage{subfigure}
\usepackage[hidelinks]{hyperref}
\usepackage{multirow}

\usepackage[noadjust]{cite}

\begin{document}

\title{Evolved neuromorphic radar-based altitude controller \\for an autonomous open-source blimp}

\author{Marina~González-Álvarez$^1$,~Julien~Dupeyroux$^1$,~Federico~Corradi$^2$,~and~Guido~C.H.E.~de~Croon$^1$
    \thanks{$^1$ Micro Air Vehicle Laboratory, Faculty of Aerospace Engineering, Delft University of Technology, Delft, The Netherlands. Contact: 
        {\tt\small g.c.h.e.decroon@tudelft.nl}}
    \thanks{$^2$ Ultra Low Power Systems for IoT, Stichting IMEC Nederland, Eindhoven, The Netherlands. }
    \thanks{This work has received funding from the ECSEL Joint Undertaking (JU) under grant agreement No. 826610 (project Comp4Drones). The JU receives support from the European Union's Horizon 2020 research and innovation program and Spain, Austria, Belgium, Czech Republic, France, Italy, Latvia, Netherlands. It has also received support from La Caixa Foundation (ID 100010434) under the code LCF/BQ/EU19/11710057.}
}

\maketitle

\begin{abstract}

Robotic airships offer significant advantages in terms of safety, mobility, and extended flight times. However, their highly restrictive weight constraints pose a major challenge regarding the available computational resources to perform the required control tasks. Neuromorphic computing stands for a promising research direction for addressing such problem. By mimicking the biological process for transferring information between neurons using spikes or impulses, spiking neural networks (SNNs) allow for low power consumption and asynchronous event-driven processing. In this paper, we propose an evolved altitude controller based on an SNN for a robotic airship which relies solely on the sensory feedback provided by an airborne radar. Starting from the design of a lightweight, low-cost, open-source airship, we also present an SNN-based controller architecture, an evolutionary framework for training the network in a simulated environment, and a control strategy for ameliorating the gap with reality. The system's performance is evaluated through real-world experiments, demonstrating the advantages of our approach by comparing it with an artificial neural network and a linear controller. The results show an accurate tracking of the altitude command with an efficient control effort.

\end{abstract}

\IEEEpeerreviewmaketitle

\section{Introduction}

\begin{figure}[t]
    \centering
    \includegraphics[width=\columnwidth]{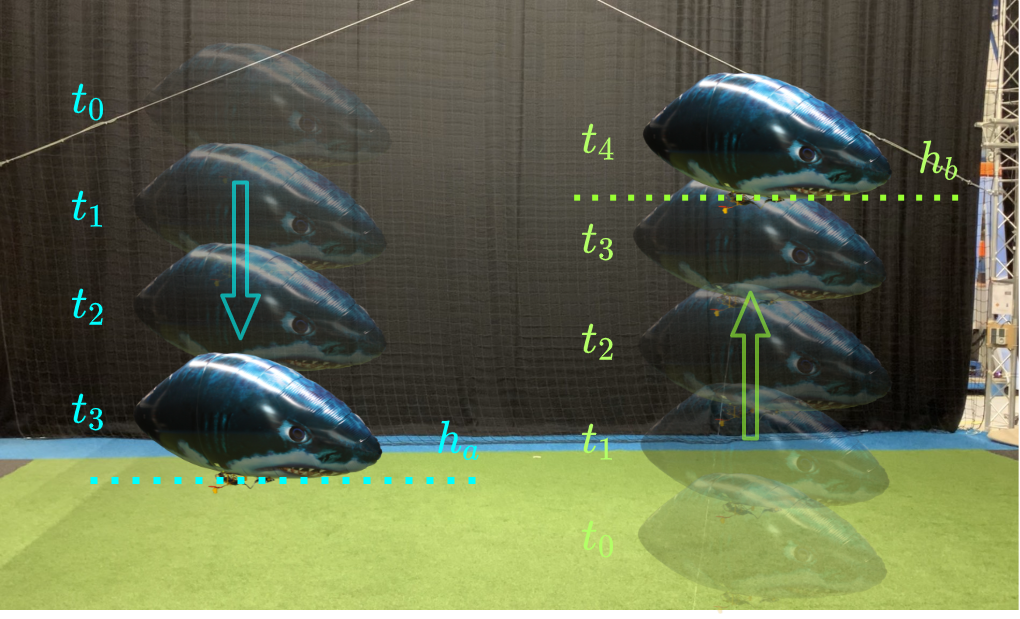}
    \caption{Proposed autonomous altitude control system for an indoor airship. Evolution of the blimp altitude for two different set-points ($h_a$ and $h_b$) for several timestamps ($t_0,\dots,t_4$).}
    \label{fig:front_figure}
\end{figure}

The biological intelligence of living beings has long attracted us to explore their innate ability to learn complex tasks. For instance, despite their limitations in terms of cognitive capabilities and energy resources, flying insects can outperform some of the most advanced aerial robots nowadays on navigating autonomously through complex environments with fast and agile maneuvers. In recent years, this inspiration has led to the development of controllers for unmanned aerial vehicles (UAVs) that mimic the structural and functional principles of the brain~\cite{Jiang2017}. Artificial neural networks (ANNs)~\cite{Abiodun2018} have proven successful for controlling different flying robots such as a hexacopter~\cite{Kusumoputro2016}, a helicopter~\cite{Suprijono2017}, or a quadrotor~\cite{Ary2017}. However, when it comes to light-weight micro air vehicles (MAVs), conventional ANNs present several disadvantages regarding energy consumption and response latency~\cite{Yousefzadeh2019}. Spiking neural networks (SNNs), are a promising research direction in this regard. By processing information using just a small population of spikes with a precise relative timing, they allow for a more efficient learning and control~\cite{Maass1997, Lee2016}. Among the main advantages of SNNs for aerial robotic applications, we can highlight that they enable computing with highly parallel architectures and provide low-power and energy-efficiency traits~\cite{loihi,ubrain}. Additionally, they are universal value function approximators~\cite{Foderaro2010b}, which theoretically makes them suitable for addressing complex control tasks. However, they have not yet become a common method for designing controllers. This is mainly due to the discrete spiking nature of SNNs, which makes the use of gradient-based optimization algorithms, such as the well-known back-propagation strategy for conventional ANNs, more challenging. To tackle these issues, in this paper we present an SNN-based altitude controller for a low-cost micro airship with an open-source design, equipped with an airborne radar (Fig.~\ref{fig:front_figure}). The choice of the problem of tracking an altitude command is motivated by its relevance in key MAV applications such as autonomous package delivery~\cite{Mathew2015, Arbanas2016}, or landing~\cite{Croon2013, Borowczyk2017}, among others. On the other hand, the selection of a lighter-than-air craft as a test platform instead of a rotorcraft, is driven by the complementary advantages it poses, such as extended flight times, excellent ease of assembly, low acoustic footprint, low power consumption, and a simpler design~\cite{Li2011}. By incorporating an airborne radar, the sensory feedback required for the control loop is robust to variant illumination and visibility conditions, while keeping the payload and computational requirements within reasonable limits.

The main contributions of this paper are twofold. First, we present an evolved altitude controller for a micro air vehicle based on an SNN, which relies solely on the sensory feedback provided by an airborne radar. We successfully demonstrate the performance of the radar-based neurocontroller onboard the aerial platform in real-world experiments, quantitatively comparing the results with those of an ANN and a proportional-integral-derivative (PID) controllers. Second, we propose the design of an open-source, low-cost, lightweight blimp platform with a 3D printable gondola, that allows for the inclusion of custom sensors and actuators. This facilitates its replication and customization for different applications. The remainder of the paper is organized as follows: Section~\ref{sec:related_work} provides an overview of the state-of-the-art in micro-airship design and spiking flight neurocontrollers. Afterwards, in Section~\ref{sec:methodology}, we present the proposed MAV design, introduce the altitude control approach based on the airborne radar, the structure of the SNN controller, the evolutionary strategy for training the network, and a blimp computational model to perform the training in a simulated environment. Then, in Section~\ref{sec:results}, we describe the real-world experimental setup as well as discuss the obtained results. Finally, Section~\ref{sec:conclusion} concludes the work and delineates future research directions.

\section{Related Work}
\label{sec:related_work}

\subsection{Micro-airship Design}

Even though the golden era of giant cargo airships has faded, the advantages offered by lighter-than-air crafts prevail. Blimps are, slowly but surely, attracting increasing interest in the realm of unmanned aerial vehicles~\cite{Artaxo2020, Price2020}. They present endless possibilities in terms of their design. For example in~\cite{Gorjup2020}, a three-propeller, low-cost platform is presented that is equipped with a camera and a compact, but closed-configuration gondola. An alternative design is proposed in~\cite{UlFerdous2019}, where the authors introduce a novel actuation mechanism based on two propellers mounted on a rotating shaft, which is oriented using a servomotor. Other examples of higher complexity include~\cite{Watanabe2015,Oh2006,Burri2013}. Although these alternatives have proven successful for their specific applications, they lack the versatility that can be achieved by leaving room for incorporating additional sensors and/or actuators. Besides, only~\cite{Gorjup2020} is open-source and lightweight enough to be mounted on commercially available blimp balloons. For the purpose of clear comparison, the main contributions of the state-of-the-art and our approach are summarized in Table~\ref{tab:blimp_soa}.

\begin{table}[h]
  \renewcommand{\arraystretch}{1.2}
  \caption{Comparison between the different blimp designs}
  \label{tab:blimp_soa}
  \centering
\resizebox{\columnwidth}{!}{\begin{tabular}{c|cccccc}
    \hline\hline
    \textbf{Property} & \textbf{\cite{Gorjup2020}} & \textbf{\cite{UlFerdous2019}} & \textbf{\cite{Watanabe2015}} & \textbf{\cite{Oh2006}} & \textbf{\cite{Burri2013}} & \textbf{Ours} \\ \hline
    Easily customizable gondola & - & \ding{51} & - & - & - & \ding{51} \\
    Low-cost design & \ding{51} & \ding{51} & \ding{51} & \ding{51} & - & \ding{51} \\
    Open-source availability & \ding{51} & - & - & - & - & \ding{51} \\
    Lightweight Microfoil blimp & \ding{51} & - & - & - & - & \ding{51} \\
    Number of propellers & 3 & 2 & 4 & 6 & 4 & 2 \\
    Number of servomotors & - & 1 & - & 3 & - & 1 \\ \hline\hline
  \end{tabular}}
\end{table}

\subsection{Spiking Neural Network-based MAV Control}

The inherent nonlinear dynamics of most MAVs makes them challenging to control. Moreover, their restrictive weight constraints inevitably limit the computational power of the controller. SNNs enable computing with highly parallel architectures made of simple integrate-and-fire neurons interconnected by weighted synapses. Implementations of spiking flight neurocontrollers include~\cite{Howard2014a}, where the authors propose an SNN for robust control of a simulated quadrotor in challenging wind conditions. They achieve a better performance in waypoint holding experiments compared with a hand-tuned PID and a multi-layer perceptron network. Another example is presented in~\cite{Clawson2016a}, where an SNN controller that adapts online to control the position and orientation of a flapping drone is proposed. SNNs have also been applied to obstacle avoidance tasks, as direct flight~\cite{Foderaro2010b} or decision-making~\cite{Zhao2018} controllers. In both cases they use reward-modulated learning rules for training the SNN. Although these MAV controllers have excelled in simulated environments, their main limitation is that they have not been evaluated in real-world experiments. The scope of works that have implemented SNN controllers for MAVs in real scenarios is much more limited. The first work that integrates an SNN in the closed-loop control of a real-world flying robot is very recent~\cite{Hagenaars2020}. There, the authors present an SNN for controlling the landing of a quadrotor by exploiting the optical flow divergence from a downward-looking camera and the readings of an inertial measurement unit (IMU). To address the learning problem of SNNs~\cite{Wang2020a}, they adopt an evolutionary training strategy. In~\cite{Dupeyroux2021}, this controller is enhanced by using hardware specifically designed for neuromorphic applications. Although not tested in free flight experiments, the potential advantages of SNN controllers implemented in these devices are also demonstrated in~\cite{Sandamirskaya2021}. Our work aims to extend the framework proposed in~\cite{Hagenaars2020}, by (1) controlling the altitude instead of landing; (2) considering an open-source micro blimp, which has less control authority and harder to model dynamics than a quadrotor; and (3) exploiting solely the range measurements provided by a radar, reducing the number of required sensors on-board.

\section{Methodology}
\label{sec:methodology}

\subsection{Open-source Micro-airship}

\begin{figure}[t]
    \centering
    \includegraphics[width=\columnwidth]{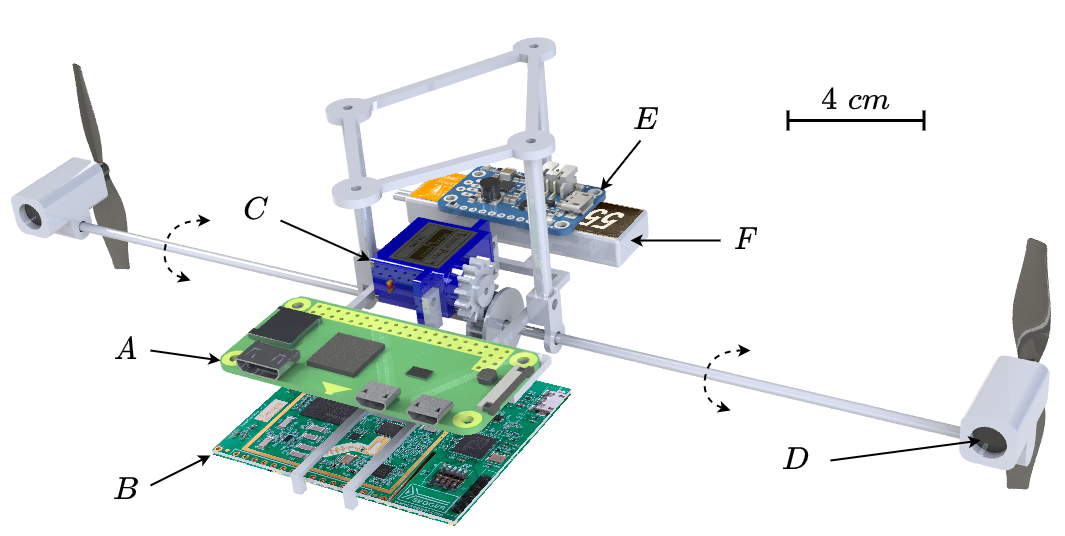}
    \caption{Proposed airship design. (A) Raspberry Pi W Zero; (B) 24 GHz Infineon Radar Position2Go; (C) Sub-micro Servo SG51R; (D) 8520 Coreless Motor; (E) PowerBoost 500 Basic; (F) 550mA 3.8V Li-Po battery.}
    \label{fig:gondola}
    \vspace{-3mm}
\end{figure}

The proposed design for the micro autonomous airship is illustrated in Fig.~\ref{fig:gondola}. The reader interested in replicating the platform can find further details, links to re-sellers, prices, and parts for 3D printing at: \url{https://github.com/tudelft/blimp_snn}. The airship's gondola can be 3D printed and assembled in a modular fashion, with a total frame weight of just $9$g. Due to its open configuration, the components mounted on the gondola can be easily interchanged, leaving room for versatility on the selection of sensors and actuators. In addition, we include a rotary shaft with a case for accommodating the propellers on both ends for controlling the altitude. Finally, we incorporate four hitches on top of the gondola, where we tape Velcro strips for attaching the envelope. Regarding the electronic components, we use a Raspberry Pi W Zero as the central communication and control unit, running the Raspbian Lite operating system. The robot's steering is achieved through the micro servomotor mounted on the gondola and the two core-less direct current (DC) motors attached at each end of the shaft. Specifically, the servo is responsible for the rotation of the shaft, up to 180\degree, and the DC motors allow for an independent control of the thrust on each side. Additional peripheral components include a step-up voltage regulator, a 500 mAh Li-Po battery and a motor driver. Finally, a fast chirp frequency-modulated continuous wave (FMCW) radar module from Infineon with a resolution of $\pm 20$ cm is used as a ranging sensor for the closed-loop control. Concerning the airship's envelope, the material chosen is Microfoil due to its excellent gas retention capabilities~\cite{Gorjup2020}. We select a model that provides the largest achievable payload among the commercially available miniature blimps ($150$g) while keeping a relatively low price. For our application, we use helium as the lifting gas. Considering all the aforementioned elements, the proposed platform weights a total of 147g. To integrate the different components and perform the computations on-board we adopt the Robot Operating System (ROS)~\cite{Quigley2009} framework. In addition, a teleoperation package to manually control the airship from a ground computer keyboard via a secure shell (SSH) connection is also provided in the repository included at the beginning of this section.

\subsection{Altitude Controllers} 

In order to control the airship's altitude, the commands are provided in terms of motor voltages, $u \in [-u_{\max},u_{\max}]$ $[V]$, with $u_{max} = 3.3$ $[V]$. The larger the absolute value of $u$, the more thrust the propellers provide, and therefore, the greater the acceleration of the blimp will be. When $u>0$, the robot moves upwards and, when $u<0$, the robot moves downwards, with the shaft rotated $180\degree$. To determine the required control actions for tracking an arbitrary reference altitude $h_{\textit{ref}}$, we process the readings from the radar to get an estimate of the current altitude $h_{\textit{curr}}$ of the blimp~\cite{Wessendorp2021} -- the range-Doppler algorithm~\cite{Winkler2007RangeDD}, along with a median filter, is used to estimate the altitude. Then, to effectively track an arbitrary altitude command, we design a controller that provides a mapping between the altitude error, $h_{\textit{ref}}-h_{\textit{curr}}$, and the motor voltages, $u$, such that the former is minimized. We consider three distinct approaches for benchmarking purposes: a linear PID, an artificial neural network, and a spiking neural network.

\subsubsection{PID controller} A conventional PID is one of the most simple, yet widespread methods for addressing control problems. In discrete form, the mapping between the error signal $e_k=h_{ref}(k)-h_{curr}(k)$ and the motor command $u_k$ is given by \cite{Ogata87}:
\begin{equation}
    u_k = K_p e_k + \frac{K_d}{T} \left(e_k - e_{k-1}\right) + K_i T \left(e_k + e_{k-1}\right)
\end{equation}
where $K_p$, $K_i$ and $K_d$ refer to the proportional, integral and derivative gains, respectively, $T$ to the sampling period, and $k$ to the timestamp. These are tuned empirically using the proposed MAV platform.

\subsubsection{ANN controller} We propose a standard, non-neuromorphic ANN controller where the tracking error $h_{\textit{ref}}-h_{\textit{curr}}$ is directly fed into the network in the form of a continuous signal. The network consists of a single neuron in the first and last layers, and two hidden layers containing 3 and 2 neurons respectively. The input and the two hidden layers operate with a $\tanh()$ activation function. At the output layer, a linear neuron provides the value of the motor command $u$, clamped to the interval $\pm u_{\max}$ $[V]$.

\begin{figure}
    \centering
    \includegraphics[width=\columnwidth]{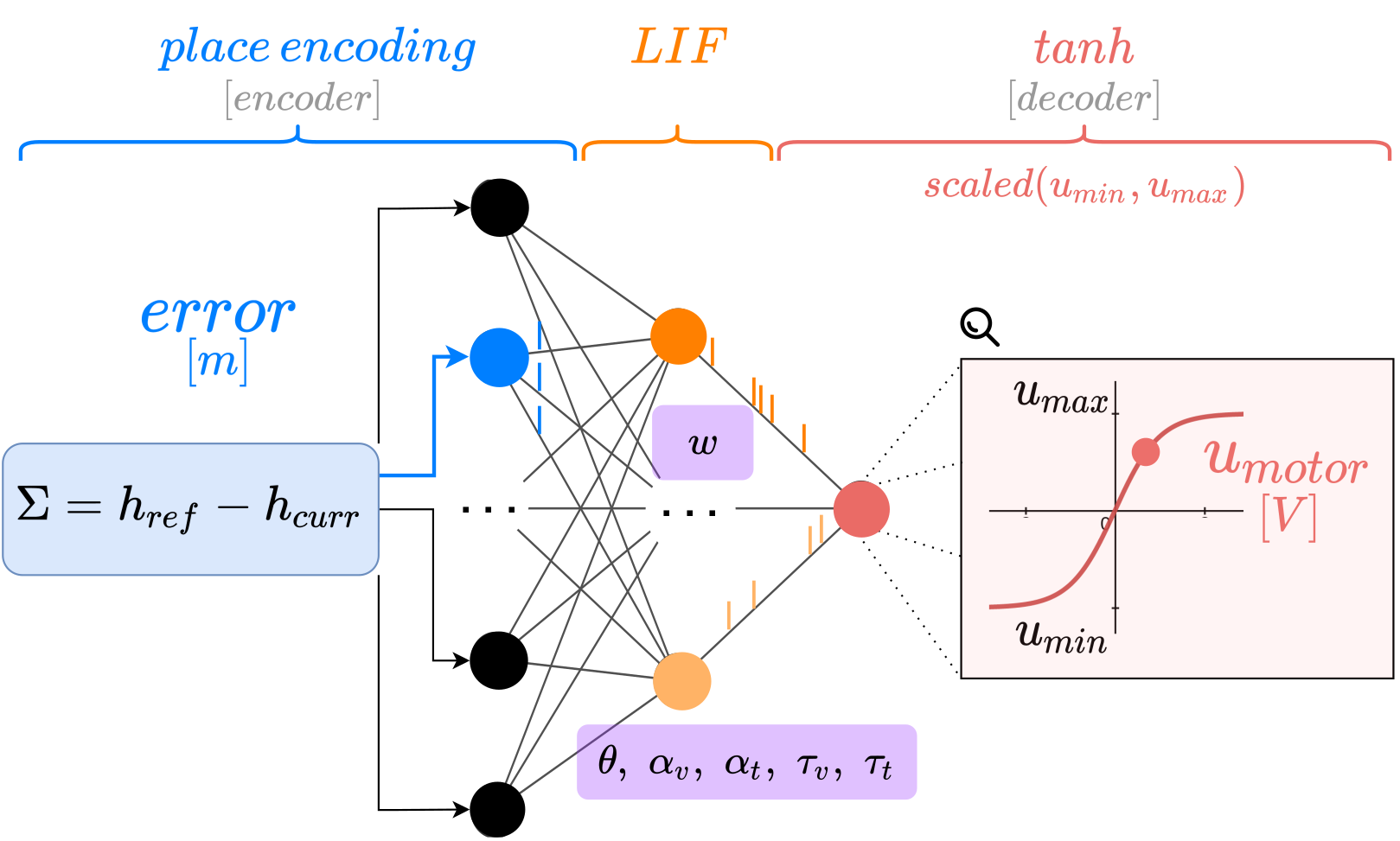}
    \caption{Schematic of the SNN controller architecture. The (evolved) network parameters are highlighted in violet, being $w$ the synaptic weights, $\theta$ the spiking threshold, $\alpha_{v/t}$ the scaling constant for the increase of the voltage/trace by a single spike, and $\tau_{v/t}$ the decay for the voltage/trace.}
    \label{fig:snn_arch}
\end{figure}

\subsubsection{SNN controller} The proposed SNN architecture is illustrated in Fig.~\ref{fig:snn_arch}: it consists of three fully connected layers (10 -- 5 -- 1 neurons). The input layer encodes the altitude error signal into spikes using position coding: the input values of $h_{\textit{ref}}-h_{\textit{curr}}$ are divided into 10 intervals, with each of them assigned to a different neuron. The first and last intervals corresponds to  $\left]-\infty,\,-0.4\right[$ and $\left]0.4,\,\infty\right[$, respectively, while the range of the remainder 8 neurons is uniformly distributed between $\left[-0.4,\,0.4\right]$. Each time the altitude error falls within one of these intervals, the corresponding neuron fires a single spike. The hidden layer consists of five leaky integrate-and-fire (LIF) neurons, where the membrane potential of the $i$-th neuron, $v_i(t)$, is governed by the following equation:
\begin{equation}
    v_i(t) = \tau_{v_i} \cdot v_i(t-\Delta t) + \alpha_{v_i} u_i(t)\qquad i=1,\dots,5
\end{equation}
referring $\tau_{v_i} \in [0,1]$ to the decay factor per time-step $\Delta t$, $\alpha_{v_i}$ to a scaling constant, and $u_i(t)$ to the synaptic input current:
\begin{equation}
    u_i(t) = \sum_{j=1}^{10} w_{ij} s_j(t).
\end{equation}
that is, multiplying the incoming spikes from the $j$-th input neuron $s_j(t)$, by the synaptic weights $w_{ij}$. Whenever the membrane potential $v_i(t)$, reaches a certain threshold $\theta_i$, a postsynaptic spike is triggered and $v_i(t)$ resets back to $0$. The output layer decodes the spikes back into a real value. It consists of a single non-spiking neuron with a scaled $\tanh()$ activation function. The neuron conducts a weighted sum of the so-called spike traces, $X_i(t)$, which acts as a low-pass filter with decay $\tau_{t_i} \in [0,1]$, and is computed as:
\begin{equation}
    X_i(t)=\tau_{t_i} \cdot X_i(t-\Delta t)+\alpha_{t_i}s_i(t),
\end{equation}
being the definition of $\tau_{t_i}$ and $\alpha_{t_i}$ analogous to $\tau_{v_i}$ and $\alpha_{v_i}$. The resulting value is scaled within the control limits, $\pm u_{\max}$. Following this, the motor command, $u$, is given by:
\begin{equation}
    u(t) = u_{\max}\cdot\tanh\left(\sum_{i=1}^5w_{i}X_i(t)\right)
\end{equation}

\subsection{Evolutionary Framework}
\label{sec:evolutionary_framework}

For training the neural controllers we adopt an evolutionary strategy with a mutation-only procedure. Each evolution begins with a randomly initialized population of $N$ individuals. The offspring is obtained by performing a randomized tournament selection of $M$ individuals, i.e., randomly selecting $M$ aspirants from the population and keeping the one with the best fitness. This is repeated $N$ times, so that the population size is invariant. The $n$-th individual is mutated with a probability of $p_{\textit{mut}}^{(n)}=0.4$, and its $m$-th parameter with $p_{\textit{mut}}^{(m)}=0.6$. These mutations take place according to uniform probability distributions $\mathcal{U}\left\{,\right\}$, whose range is shown in Table~\ref{tab:nn_params} for both the SNN and ANN. For the latter, the open parameters are the biases, $b_i$, and analogously to SNNs, the weights, $w_{ij}$.

\vspace{-1mm}

\begin{table}[h]
    \renewcommand{\arraystretch}{1.2}
    \centering
    \caption{Parameters mutated during evolution}
    \vspace{-1mm}
    \label{tab:nn_params}
    \begin{tabular}{c|ccc}
         \hline\hline
         & \textbf{Parameter} & \textbf{Domain} & \textbf{Mutation} \\ \hline
         \multirow{3}{*}{\textbf{SNN}} & $w_{ij}$ & $[-5,\dots,5]$ & $\mathcal{U}\left\{-2.5,2.5\right\}$ \\
         & $\theta_{i}$, $\tau_{{v_i}}$, $\tau_{{t_i}}$ & $[0,\dots,1]$ & $\mathcal{U}\left\{-0.5,0.5\right\}$ \\
         &  $\alpha_{v_i}$, $\alpha_{{t_i}}$ & $[0,\dots,2]$ & $\mathcal{U}\left\{-1.0,1.0\right\}$ \\ \hline
         \textbf{ANN} & $w_{ij}$, $b_{i}$ & $[-5,\dots,5]$ & $\mathcal{U}\left\{-2.5,2.5\right\}$ \\ \hline\hline
    \end{tabular}
\end{table}

The mutated offspring is then evaluated in a model-based simulation environment (see Section \ref{subsec:model-based_sim}), where a source of random Gaussian noise is added to the radar signal. Since this randomization stimulates the persistence of controllers that are independent of such disturbances, it helps minimizing the reality gap~\cite{Scheper2020}. During the evaluation, a set of $10$ different reference altitudes $h_{\textit{ref}}\in [0,3]$ is provided along a total simulated duration of $T = 15$ seconds each. The fitness of each individual is then quantified as the root mean squared altitude error (RMSAE):
\begin{equation}
    \label{eq5}
    \text{RMSAE} = \sqrt{\frac{1}{T}\sum_{k = 0}^{T}  \left( h_{\textit{ref}}(k) - h_{\textit{curr}}(k) \right)^2}
\end{equation}

During the evolution process, a \textit{hall of fame} which holds the 5 best performing individuals across all generations, is maintained. This prevents discarding those who have achieved a good performance. After $N_{\textit{gen}}$ generations, the individuals are also reevaluated on five more random sets of altitudes to increase the robustness. The best-performing ones are selected for further real-world experiments. The architecture of the neural controllers is kept fixed during the evolution. The choice for a specific network architecture is made experimentally, after repeating this training process several times for networks of different complexities. Concretely, the simplest and smallest network which does not compromise the control performance is selected.

\subsection{Model-based Simulation Environment}
\label{subsec:model-based_sim}

The altitude controllers evolve in a simulated environment since it would be infeasible to perform all the required evaluations in the real world. For that, we develop a dynamical model of the blimp to obtain a mapping between the motor commands and the evolution of the blimp's altitude over time. To simplify the training and avoid adding further complexity, we consider a linear model. We assume that the acceleration at the $k$-th time step $\ddot{h}_k$, is proportional to the voltage applied to the motors, $u$, i.e.
\begin{equation}
    \ddot{h}_k = a_1 u_{k-1} + a_2 u_{k-2}
\end{equation}
where $a_i$ is the proportionality constant for the motor command at time instant $k-i$. However, since the acceleration cannot be directly measured with the radar sensor, we can instead express this relation in terms of the measured altitude, $h$ by taking Euler's discretization of the derivative. Applying the Z-transform, we obtain the following transfer function, which maps the commands $u_k$ to the altitude $h_k$:

\begin{equation}
    \label{eq:blimp}
    h_k = \frac{a_1z^{-1}+a_2z^{-2}}{1-2z^{-1}+z^{-2}} u_k
\end{equation}

To determine the unknown parameters $a_i$, we collect a data-set by tele-operating the blimp and measuring its altitude over time. After subtracting the mean, we infer the model parameters by minimizing the normalized root mean squared altitude error (NRMSAE), which can be interpreted as a measure of how well the expected response $h_k$ matches the observed data $\hat{h}_k$.

\section{Results}
\label{sec:results}

\subsection{Experimental Setup}

\subsubsection{Simulation} To train the neural controllers, we evolved five randomly initialized populations of 100 individuals through 300 generations, following the procedure described in Section \ref{sec:evolutionary_framework}. The implementation of the evolutionary optimization is based on the Distributed Evolutionary Algorithms in Python (DEAP)~\cite{deap2012} framework, while the simulation of the networks is performed by means of the PySNN library \cite{BasBuller2019}.

\subsubsection{Real-World} 
\begin{figure}
    \centering
    \includegraphics[width=\linewidth]{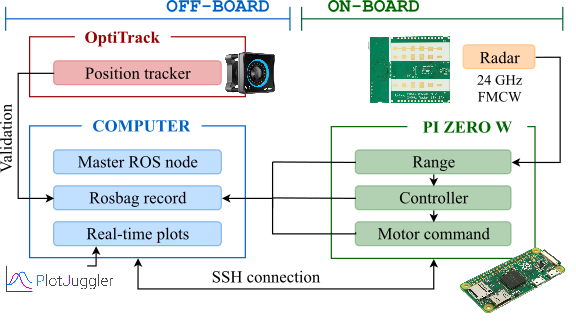}
    \caption{System overview. After processing the feedback provided by the radar sensor, an estimate of the range is sent to the Raspberry Pi Zero W control unit. Only data recording and real-time plotting operations are conducted on the ground computer, which communicates with the Pi via an SSH connection. The OptiTrack is used during the post-processing stage just for validation purposes.}
    \label{fig:real_all}
\end{figure}

An overview of the setup is shown in Fig.~\ref{fig:real_all}. The on-board control unit is a 1GHz single-core processor Raspberry Pi Zero W with 512MB RAM. The Infineon Position2Go radar provides altitude measurements. The control loop runs at a rate of 5 Hz.

\subsection{Blimp Model}

\begin{figure}
    \centering
    \includegraphics[width=0.85\columnwidth]{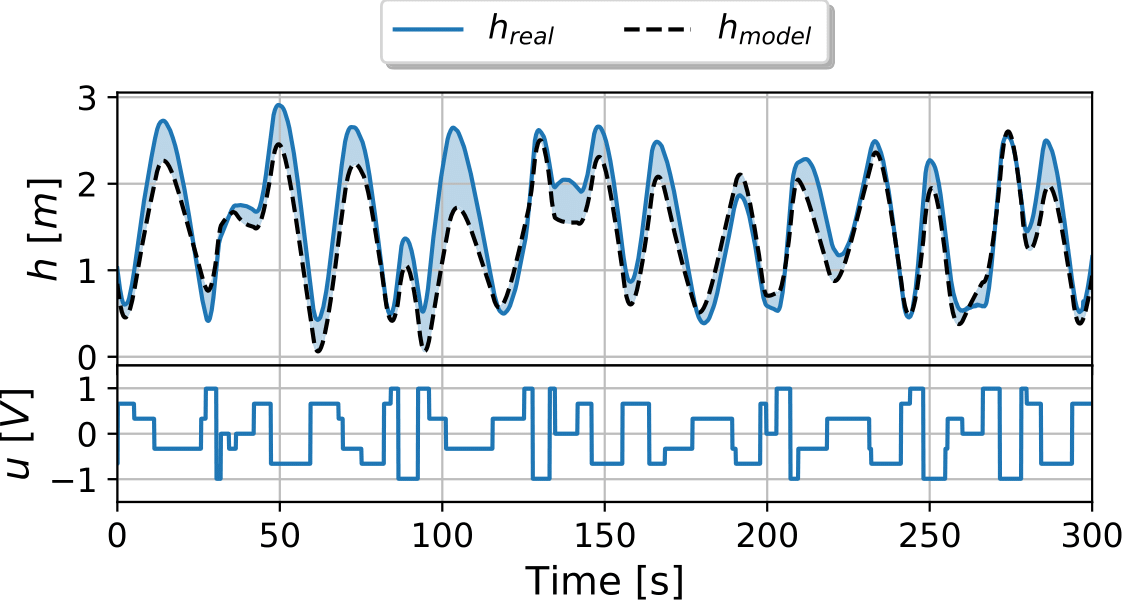}
    \caption{Validation of the blimp model. \textbf{\textit{Bottom:}} Motor commands. \textbf{\textit{Top:}} The ground truth evolution of the altitude, $h_{\textit{real}}$, compared with the evolution predicted by the model, $h_{\textit{model}}$. The error is represented by the blue area.}
    \label{fig:results_model}
\end{figure}

Following the procedure explained in Section~\ref{subsec:model-based_sim}, we infer the parameters of a blimp model of the form~\eqref{eq:blimp}, based on experimental data gathered using the real hardware. In Fig.~\ref{fig:results_model} we show a comparison between the evolution of the altitude predicted by the model and the ground truth when applying identical motor commands. We can see that we are able to reproduce the blimp's behavior using the proposed data-driven model, with a RMSAE of 0.27m over the 300 seconds run.

\subsection{Controller Evaluation}
\label{subsec:sim_real_results}

We evaluate the performance of three different altitude controllers based on a linear PID, an ANN, and an SNN. The tracking precision is tested on a sequence of five different way-points $h_d=\left\{3,2,1,2.5,1.5\right\}$m, maintained during $60$s.

\subsubsection{PID controller} 
\begin{figure}[t]
    \centering
    \includegraphics[width=0.87\columnwidth]{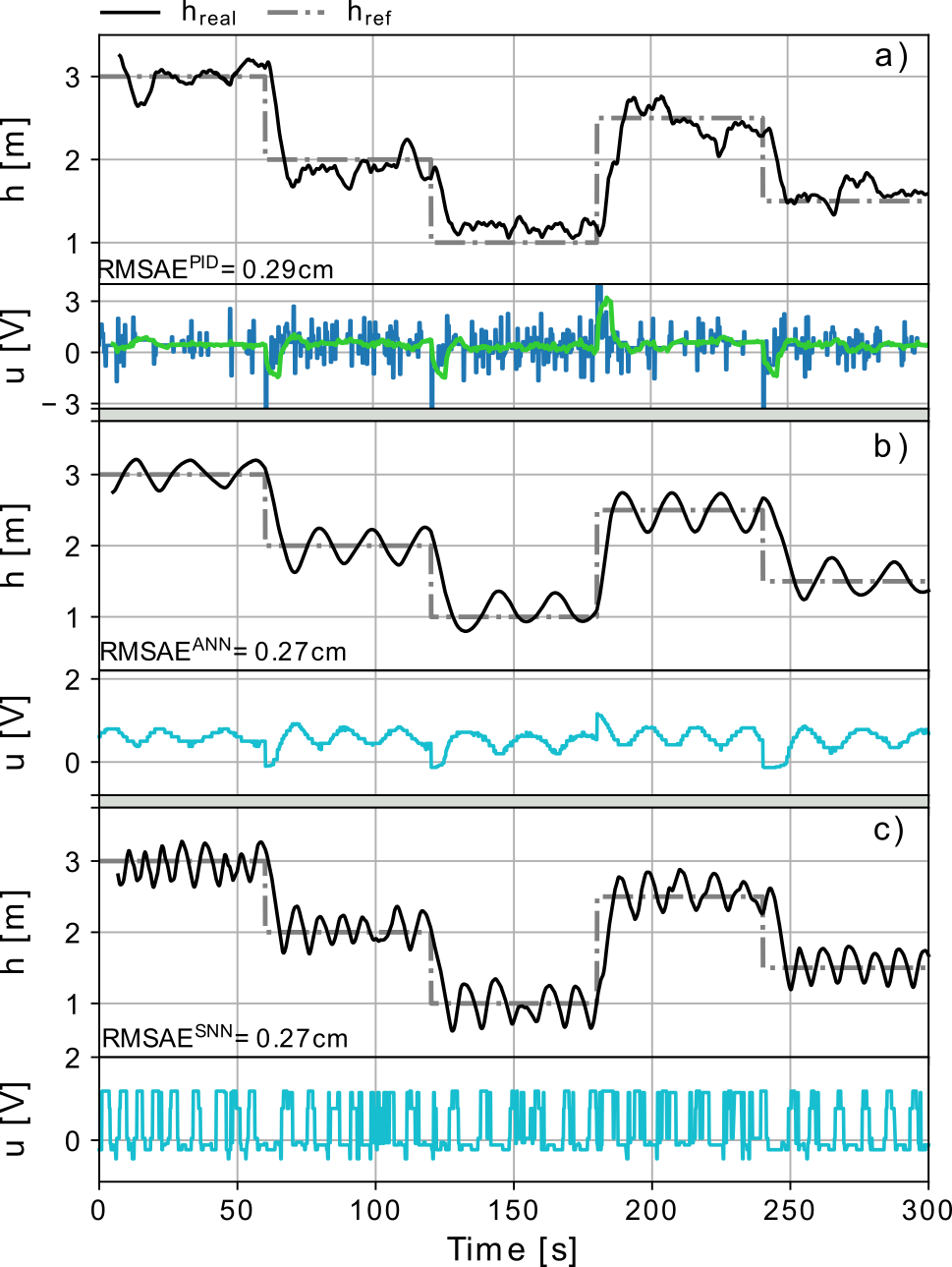}
    \caption{Experimental evaluation of the considered controllers. For all three sub-figures, at the bottom we have the motor commands, and on top, the evolution of the blimp's altitude $h_{\textit{real}}$ compared with the reference $h_{\textit{ref}}$. (a) PID. (b) ANN. (c) SNN.} 
    \label{fig:results_ann_and_snn}
\end{figure}

The experimental results are depicted in Fig.~\ref{fig:results_ann_and_snn}(a). We can see that we can track the altitude commands effectively. Quantitatively, we obtain a RMSAE of 0.29m, which indicates a satisfactory performance, considering that the uncertainty of the radar sensor is of $\pm0.2$m.

\subsubsection{ANN controller} The obtained results are shown in Fig.~\ref{fig:results_ann_and_snn}(b). We can see that the blimp effectively converges to the altitude set-point but presents an oscillatory behavior. This is mainly because of two reasons: the minor contribution of the diminished discrepancies between the model and the vehicle's inherent dynamics; and the slow responsiveness of the system, especially when the motor commands are not too abrupt, as it is the case. However, it can be noted that the trajectory is smoother than with a PID. The RMSAE now corresponds to $0.27$m.

\subsubsection{SNN controller} The experimental results for this case are displayed in Fig.~\ref{fig:results_ann_and_snn}(c). We note that the behavior and performance are similar to the ANN case, but with faster oscillations, due to the output's binary nature caused by the presence or absence of the spike. The RMSAE is of $0.27$m.

\section{Conclusion}
\label{sec:conclusion}

In this paper, we aim at paving the way toward fully neuromorphic control for flying vehicles. The results obtained in real-world experiments successfully demonstrate the system's performance. By comparing it with a standard PID and an artificial neural network, we corroborate the advantages offered by SNNs in terms of adaptability and low control effort -- ultimately, the use of SNNs on-board neuromorphic controllers such as the Loihi~\cite{Davies2018} will fully demonstrate their advantages in terms of processing time and energy, while helping closing the loop towards a fully neuromorphic control of robotic systems.

\end{document}